\documentclass{bmvc2k}

\title{A data augmentation methodology for training machine/deep learning gait recognition algorithms}

\addauthor{Christoforos C. Charalambous}{http://www.bicv.org/team/christoforos-charalambous/}{1}
\addauthor{Anil A. Bharath}{http://www.bicv.org/team/anil-anthony-bharath/}{1}

\addinstitution{
 \qquad \quad \quad BICV Research Group\\ \vspace{1mm}
 \qquad \quad \quad Department of Bioengineering\\
 \qquad \quad \quad Imperial College London\\
 \qquad \quad \quad London, UK
}

\runninghead{C. Charalambous, A. Bharath}{Data augmentation for gait recognition}

\begin{document}

\maketitle

\begin{abstract}
There are several confounding factors that can reduce the accuracy of gait recognition systems. These factors can reduce the distinctiveness, or alter the features used to characterise gait; they include variations in clothing, lighting, pose and environment, such as the walking surface. Full invariance to all confounding factors is challenging in the absence of high-quality labelled training data. We introduce a simulation-based methodology and a subject-specific dataset which can be used for generating synthetic video frames and sequences for data augmentation. With this methodology, we generated a multi-modal dataset. In addition, we supply simulation files that provide the ability to simultaneously sample from several confounding variables. The basis of the data is real motion capture data of subjects walking and running on a treadmill at different speeds. Results from gait recognition experiments suggest that information about the identity of subjects is retained within synthetically generated examples. The dataset and methodology allow studies into fully-invariant identity recognition spanning a far greater number of observation conditions than would otherwise be possible. 

\end{abstract}

\section{Introduction}
\label{sec:intro}

\vspace{-0.25cm}

Gait has received significant attention as a biometric, particularly with the increasing quality of security cameras, the growth in city populations and the subsequent interest in tracking the movement of people in such environments. There have been several studies \cite{Bianchi,Murray} that support the usefulness of gait for identity recognition within a sample population. However, there are many confounding factors that can reduce usefulness of gait as an identifying biometric; these range from the difficulties in extracting particular signatures to the distinctiveness and robustness of those features to confounding factors. The confounding factors can be divided into three different categories, according to whether they are related to:
\begin{itemize}
	\item{the environment in which the subject is being captured (e.g. weather conditions, walking surface, background, ambient light, shadows, occlusions by other objects)}
	\item{the imaging system used in capturing the subject (e.g. subject-camera relative viewing angle, lighting conditions, camera lens distortions, image modality)}
	\item{the appearance of the subject (e.g. clothing, footwear, speed, gender, age, physical state, injuries)}
\end{itemize}

Emerging gait recognition systems depend on supervised machine/deep learning. Machine learning algorithms (especially deep networks) require vast amounts of application-specific, high-quality labelled training data, which is either very expensive or not feasible to acquire. In order to collect sufficient amounts of labelled data for training deep networks for gait recognition from scratch, participants would need to walk for hours. In addition, to collect data under different conditions, the experiment would have to be repeated multiple times, to cover all the possible combinations of conditions that are likely to occur. The only alternative strategy would be to employ transfer learning, but it is hard to find a problem for transfer learning which is sufficiently close to the complexity of gait recognition.

Taking several confounding variables together poses significant challenges. How does one gather sufficient variations across all confounding factors, and with many different subjects, within a finite time in order to train, let alone test, a system to provide accurate recognition? One strategy would be to synthesize gait motion within virtual environments in which clothing, lighting, camera pose and environments can be systematically changed. This paper takes a first step towards showing that this is feasible, conditional on sufficient quality of real motion capture (mocap) data, rendering and imaging models.

We suggest a method to generate augmented labelled training data for machine learning, including quantities of data that could be used for deep supervised learning algorithms. A new multi-modal dataset is introduced, consisting of more than 6.5 million frames of real motion capture data, video data and 3D mesh models. Based on the motion capture data collected from 26 subjects, we also provide simulation files with several confounding factors being controllable (see Figure \ref{fig:covFactors}). The motion data itself is captured from real people walking and running on a treadmill at different speeds.

Generating synthetic data for data augmentation is not new in the field of Computer Vision. Studies like \cite{Krizhevsky,Shotton,Sandwell} use synthetic generated data to augment training and/or testing sets, while others use it to learn features invariant to certain conditions \cite{Zhang}. While such methods focus on a very limited number of covariate factors, to our knowledge, this is the first attempt to provide the ability to generate synthetic video data with so many controllable conditions from real human motion data. This opens the possibility of bootstrapping or pre-training gait recognition methods that can identify people by their gait while being invariant to several different confounding factors simultaneously. We discuss the process by which this dataset is generated and demonstrate that characteristics of identity are preserved within the motion of the synthetically generated data.

\section{Literature Review}
\label{sec:litRev}


This work focuses on generating subject-specific synthetic video data for the purpose of data augmentation. Data augmentation provides a means for increasing the quantity of training data available for machine learning, and is particularly relevant when training deep learning systems from scratch \cite{Krizhevsky}.


In the context of gait recognition, data augmentation is likely to be useful for appearance-based, rather than model-based, approaches. The key idea behind appearance-based gait recognition is to collapse temporal information, from a video or a series of frames, down to a single image or a pair of images. An early attempt \cite{Bobick} to analyse human motion, suggested using two complementary images, the (binary) Motion Energy Image (MEI) and the (scalar valued) Motion History Image (MHI). Inspired by \cite{Bobick}, Han introduced the Gait Energy Image (GEI) \cite{Han}, a grayscale image, calculated by averaging the binary silhouettes of a person walking across the duration of a complete gait cycle. A different gait representation, the Key Fourier Descriptors (KFDs), proposed by Yu et al \cite{Yu}, used frequency information of the contours of binary silhouettes to extract gait features. Subsequent work by Lam et al. introduced the Motion Silhouette Image (MSI) \cite{Lam}. Inspired by the MHI, the MSI is a grayscale image created from binary silhouette sequences of ambulatory motion. The intensity value of a pixel in a MSI is a non-linear function of the temporal history of motion at that pixel location.

An alternative approach by Liu et al. \cite{Liu} suggested the Gait History Image (GHI), which addresses some shortcomings of the MEI, MHI and GEI metrics. A slightly different approach by Bashir et al \cite{Bashir} took advantage of the ability of Information Entropy to calculate the average amount of information in a signal; they introduced the Gait Entropy Image (GEnI), which is obtained by calculating the Shannon Entropy for each pixel in a sequence of size-normalised, and centre-aligned, binary silhouettes.

Most of the appearance-based methods that collapse spatio-temporal information down to single images show promise in discriminating subjects when applied under controlled imaging environments. When used in less controlled environments, the many factors that act as confounding variables can substantially reduce recognition performance. The most common confounding factor is the camera-subject relative viewing angle. One of the studies aimed at providing viewpoint-invariance in gait recognition is the work of Makihara et al. \cite{Makihara}.  The authors suggested a View Transformation Model (VTM), which is applied to Fourier-based features, transforming the features from one camera-subject viewpoint relationship into features from one or more reference views. The authors construct a feature matrix that has a different row for each viewpoint and a different column for each subject. This feature matrix is then decomposed using Singular Value Decomposition (SVD) and, by utilising simple algebra, analytical expressions to estimate the feature vector in a specific viewpoint are derived. 

Perhaps a greater challenge for gait recognition algorithms is to deal with different clothing. Nandy et al. \cite{Nandy} conducted a recent study into the possibility of clothing-invariant subject recognition. Based on the GEnI \cite{Bashir}, the authors built a new feature vector, considering the width of each row in the calculated GEnI. Their results suggest that this feature extraction process not only reduces the feature space dramatically, but also includes distinctive signatures while neglecting redundant static information. The reliability of the proposed feature was evaluated using statistical tests, such as pair-wise clothing correlation and intra-clothing variance.

In real-world scenarios it is also common to encounter occlusions due to objects between the camera and the human subject. Hofmann et al. \cite{Hofmann} attempted to address the problem of dynamic and static inter-object occlusion, introducing a new dataset and baseline algorithms that used colour histograms and the GEI feature image \cite{Han}. Another very common confounding factor, which directly affects appearance-based gait recognition algorithms, is speed of walking. Ghuan et al. \cite{Ghuan13}, employ Gabor functions from five scales and eight orientations to generate the Gabor-GEI feature template. The generated Gabor-GEI features were projected to subspaces using a Random Subspace Method (RSM), followed by 2D Linear Discriminant Analysis (2DLDA), which then projected the samples into a canonical space. The authors aimed to reduce the generalisation error by using a large number of random subspaces/base classifiers.



Data augmentation has been widely used in some areas of computer vision that rely heavily on machine learning, where suitable augmentation can help achieve object-camera pose invariance. Common forms of image augmentation include simple transformations such as left-right flipping, rotations, translations or scaling of an image \cite{Wang,Krizhevsky}. Flipping, cropping and resizing have also been used  in gait recognition \cite{Han}. However, this represents only a small subset of transformations to which we wish a gait recognition system to be invariant.

\section{Data Collection}
\label{sec:dataCollection}


Given that the objective of this work is to support the ability to generate augmented gait sequences in such a way as to incorporate controllable confounding factors, we sought to use a combination of computer generated imagery, driven by real motion capture data of human volunteers. The motion of these subjects was captured whilst they walked and ran on a treadmill at five different walking speeds (3-7 km/h) and at five different jogging/running speeds (8-12 km/h).

\begin{figure}[h]
\begin{tabular}{cccccc}
\bmvaHangBox{\fbox{\includegraphics[width=0.1125\textwidth]{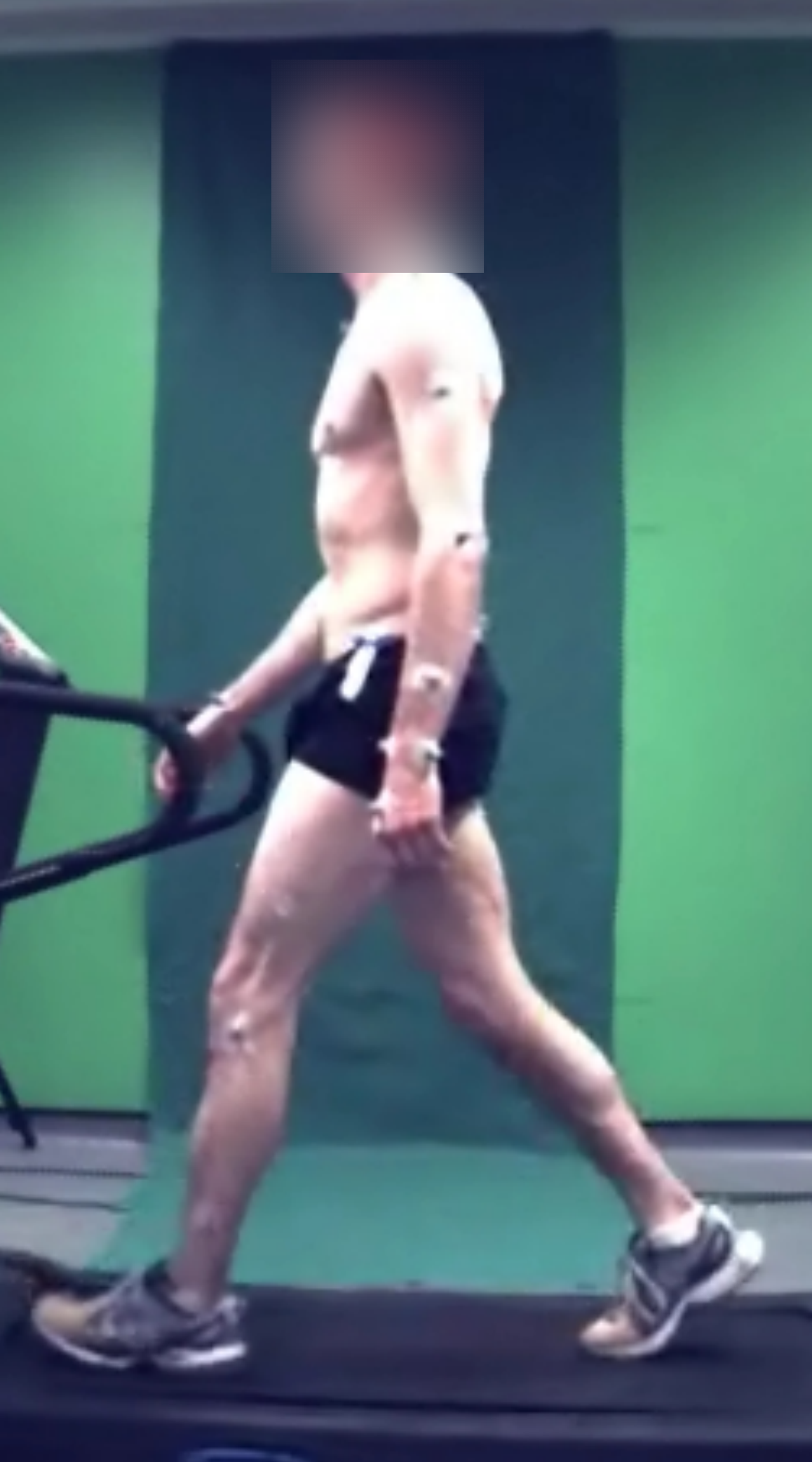}}}&
\bmvaHangBox{\fbox{\includegraphics[width=0.11675\textwidth]{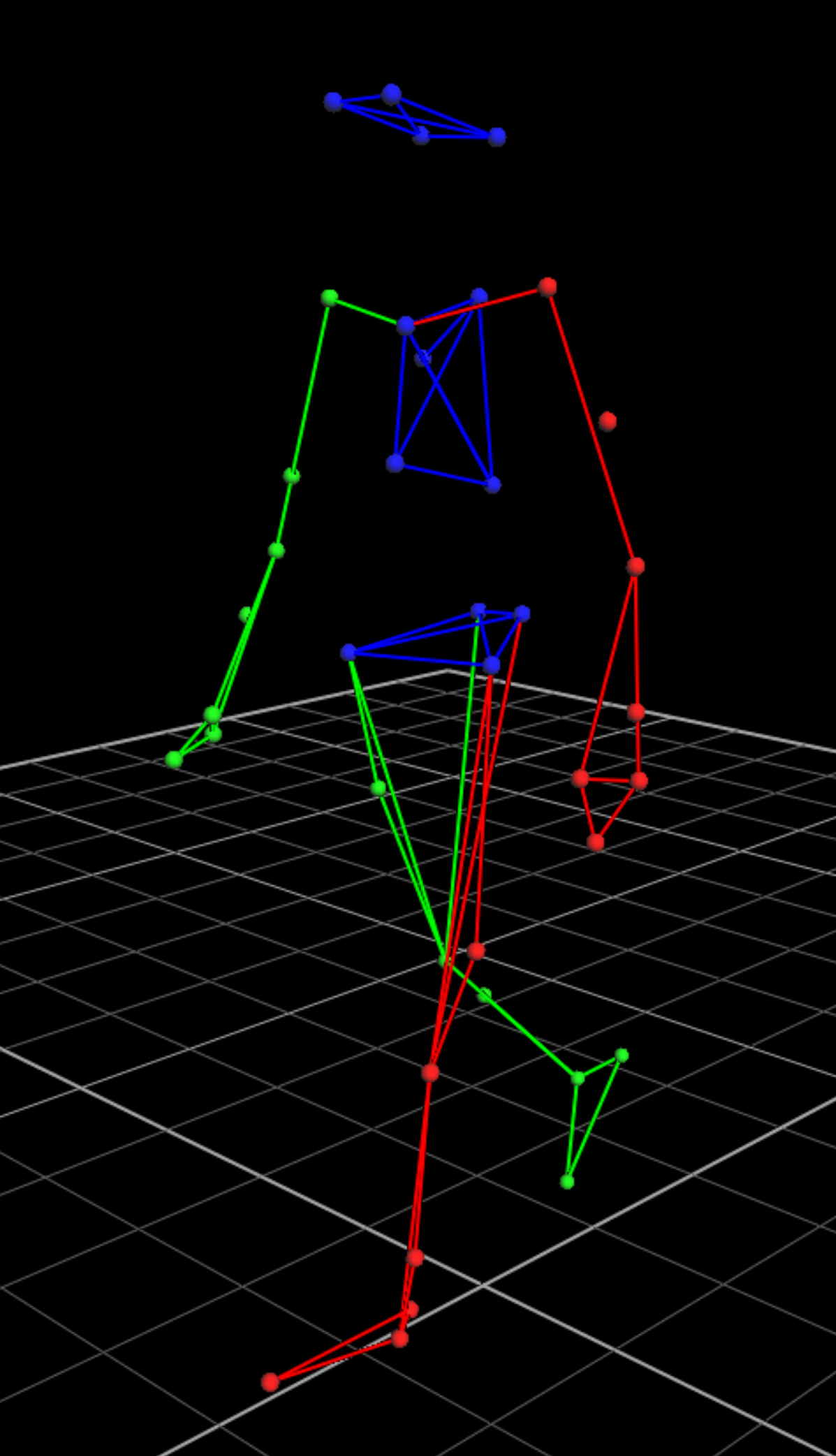}}}&
\bmvaHangBox{\fbox{\includegraphics[width=0.11675\textwidth]{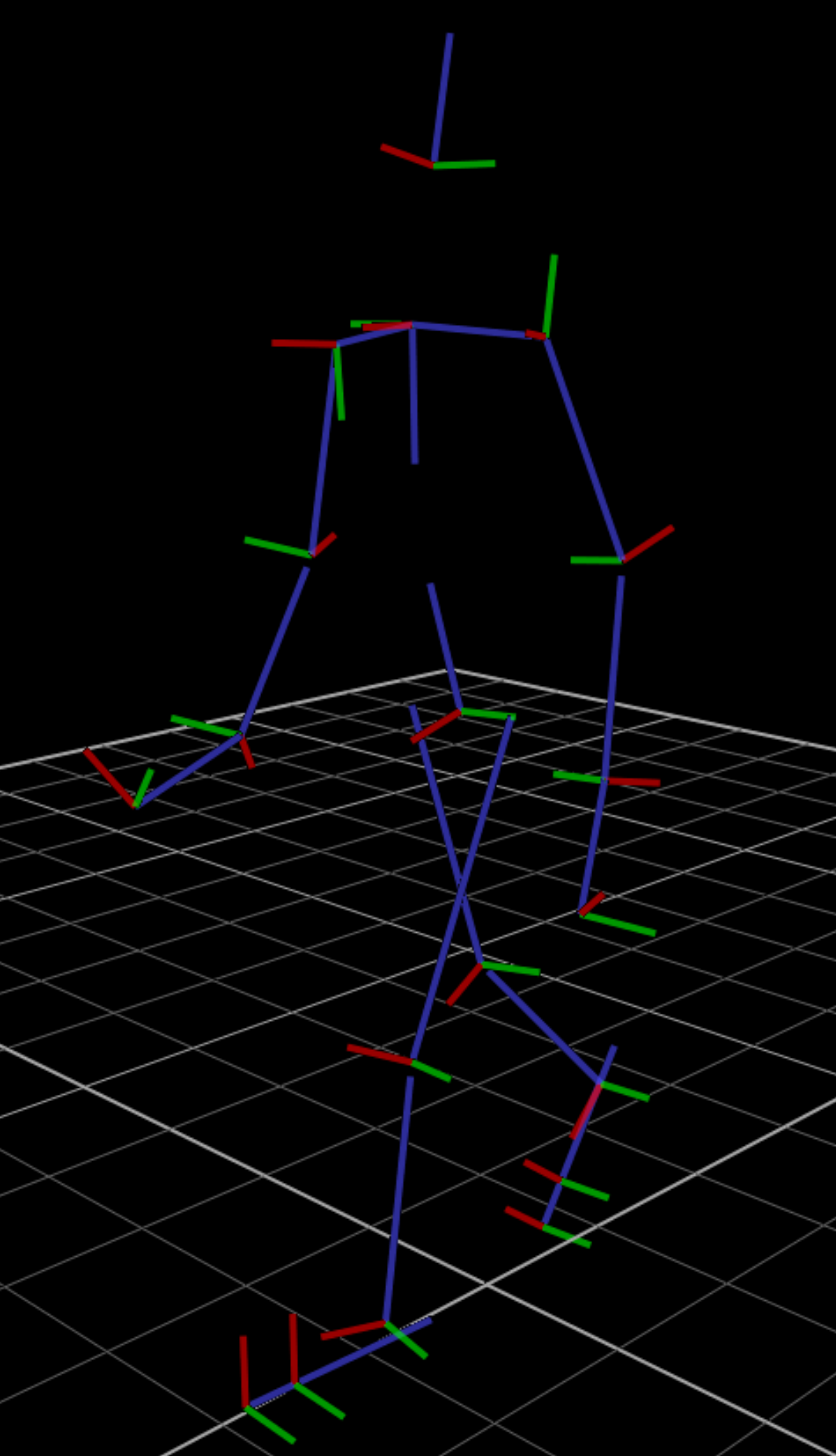}}}&
\bmvaHangBox{\fbox{\includegraphics[width=0.1128\textwidth]{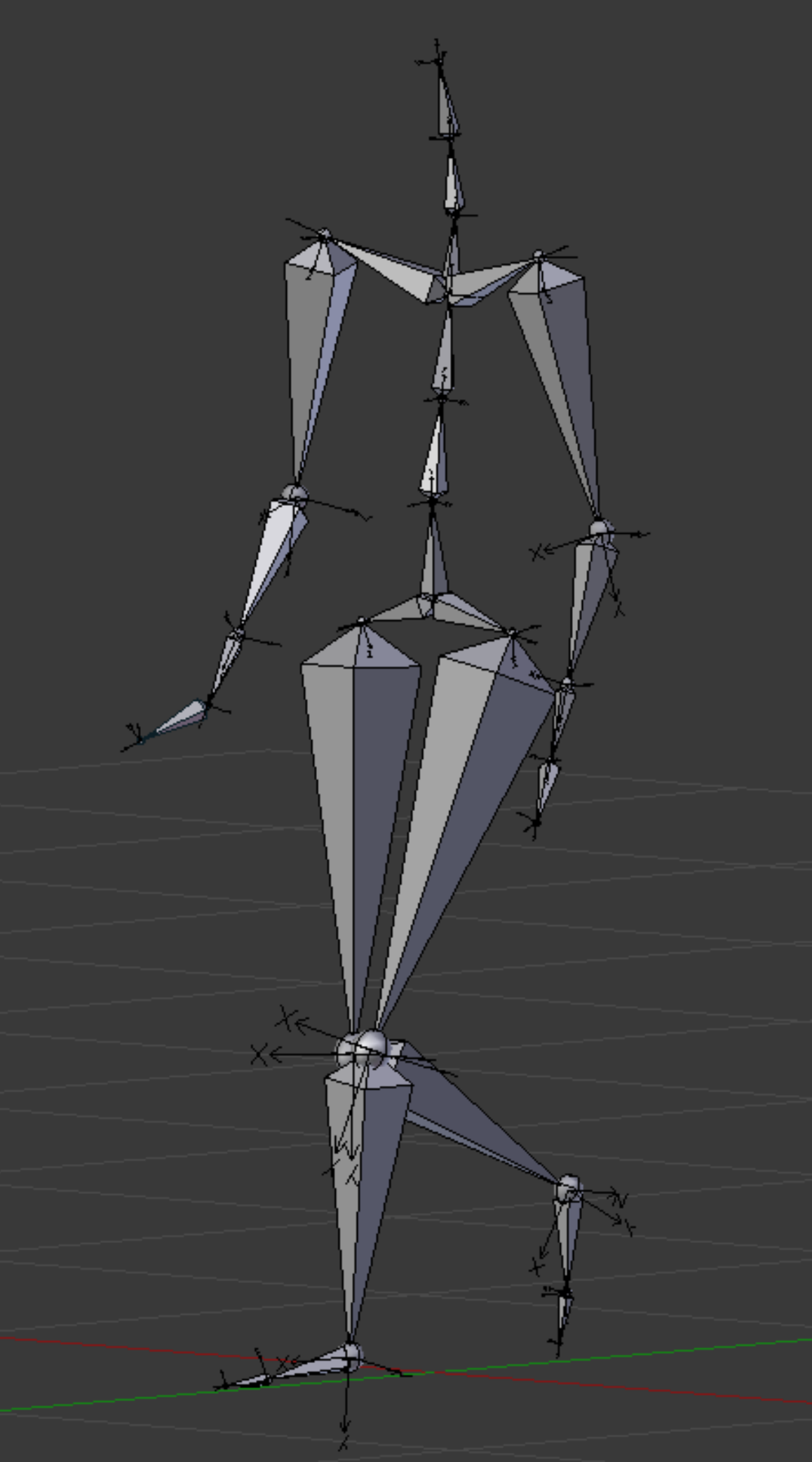}}}&
\bmvaHangBox{\fbox{\includegraphics[width=0.1233\textwidth]{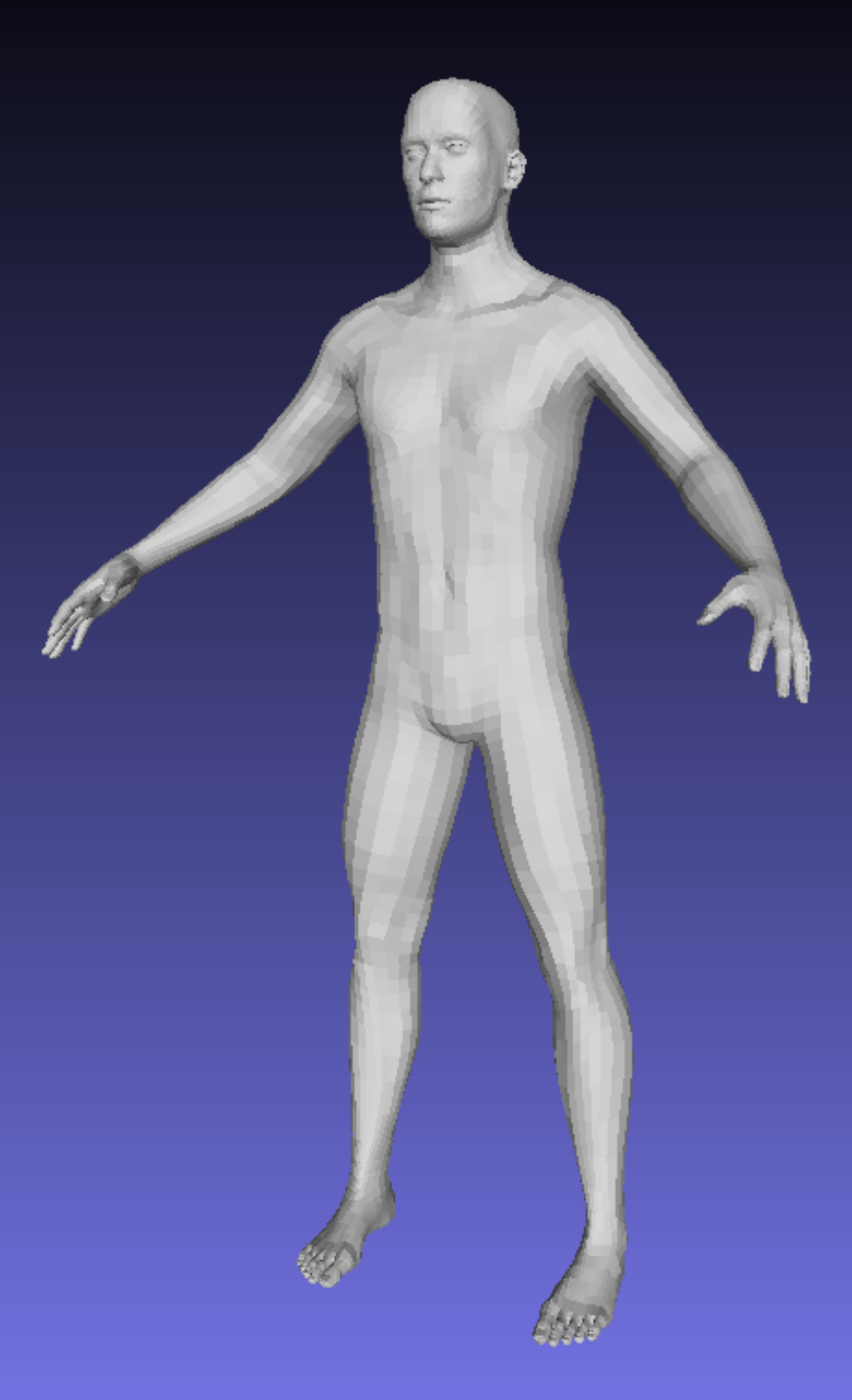}}}&
\bmvaHangBox{\fbox{\includegraphics[width=0.1125\textwidth]{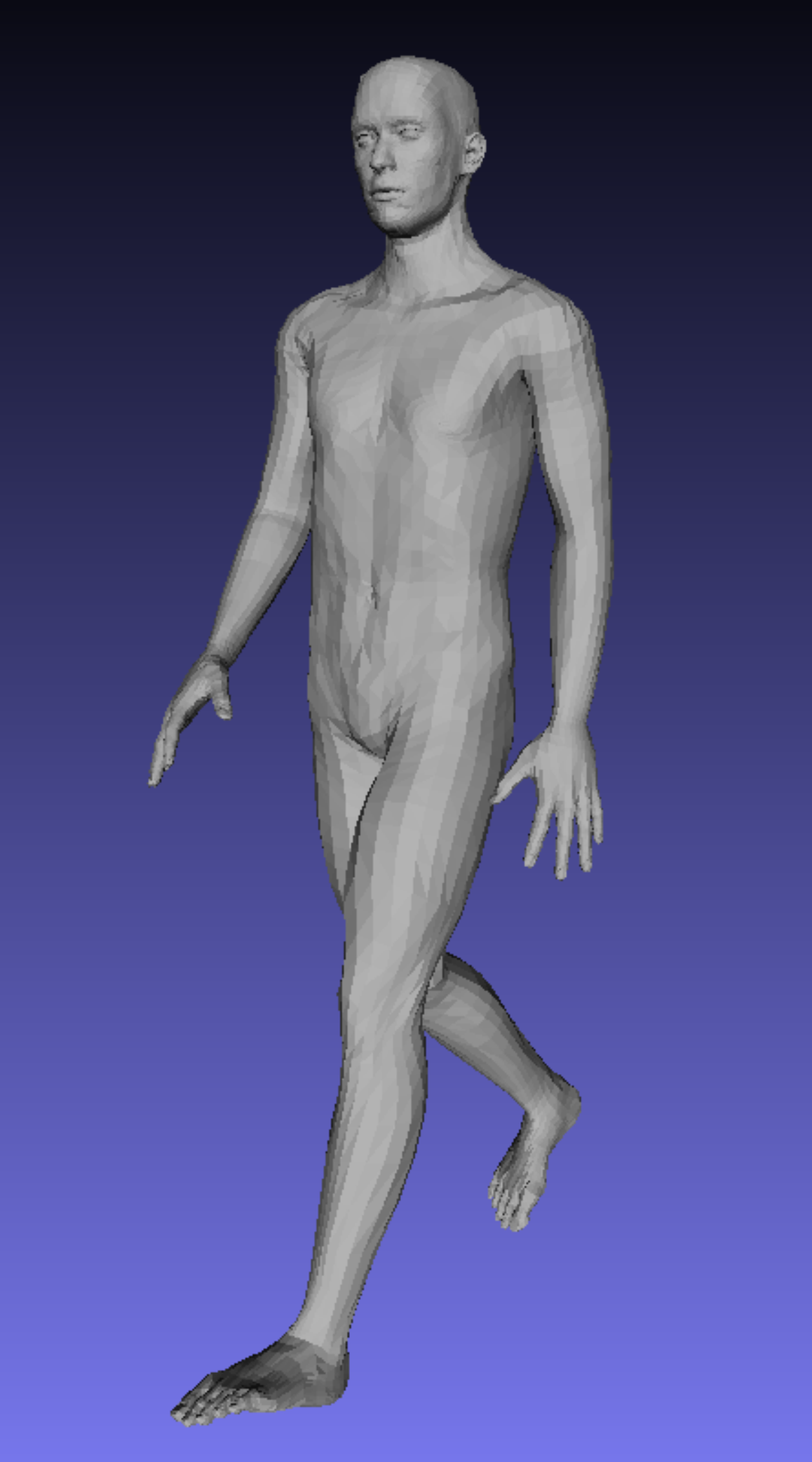}}}\\
(a)&(b)&(c)&(d)&(e)&(f)
\end{tabular}
   \caption{Selected screenshots showing key steps of the overall pipeline, for subject 3. (a) the subject with reflective markers attached to the body, (b) the 3D representation of the markers in Vicon\texttrademark\: Nexus software, (c) the estimated 3D skeleton model, (d) a skeleton following the real mocap data within Blender, (e) the 3D mesh model created in the MakeHuman environment, and (f) the 3D mesh deformed according to the collected motion capture data.}
\label{fig:pipeline}
\end{figure}

The motion capture data was collected using a multi-camera Vicon\texttrademark\: motion capture system, operating at 200 Hz. The system was configured so that 10 infrared cameras were strategically spaced around the treadmill, which was placed in the middle of the lab. A total of 38 highly reflective markers were attached on selected landmarks on the body of each subject, as seen in Figure \ref{fig:pipeline}(a). Figure \ref{fig:pipeline}(b) shows the 3D representation of the markers attached on the same subject. The 3D coordinates of the markers, along with a number of anthropometric measurements that are taken for each subject at the start of the experiment, provide the necessary information to fit a 3D skeleton model to the subject (Figure \ref{fig:pipeline}(c)). The estimated 3D skeleton consists of a total of 19 joints, fully characterised by their 3D position and orientation. Vicon\texttrademark\: Plug-in-Gait provides additional estimated kinetics (i.e. forces and moments), which are also available. However, they are not used in this work. The Vicon\texttrademark\: Nexus software \cite{Nexus}, was used to pre-process the collected data, and to estimate the 3D skeleton.

%

\section{Data Preparation}
\label{sec:dataPrep}

\begin{figure*}[h]
\begin{center}
\bmvaHangBox{\fbox{\includegraphics[width=0.55\textwidth]{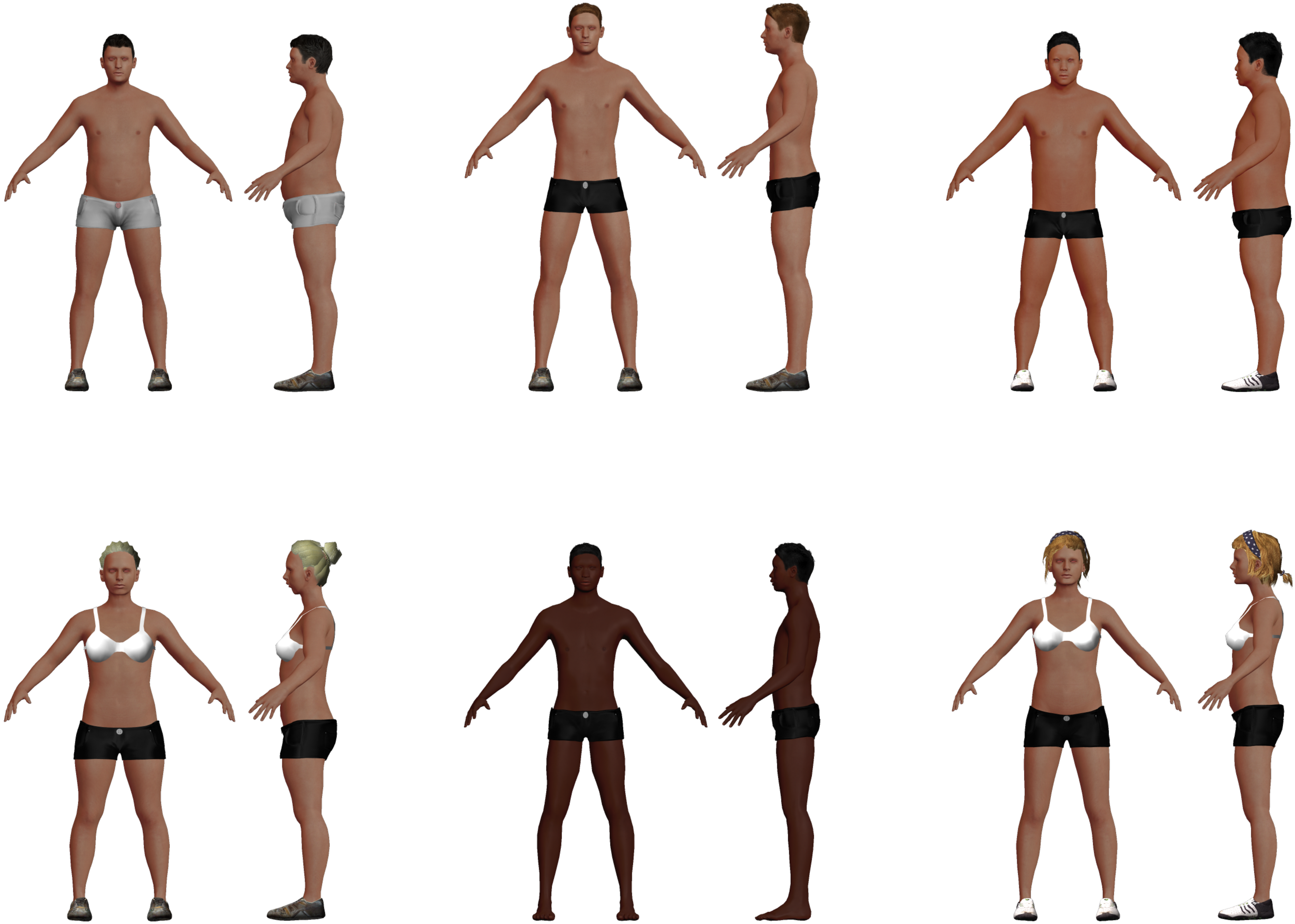}}}
\end{center}
   \caption{Front and side views of the avatars created for six different human subjects from our dataset (starting from upper left, going clockwise) subjects 1,3,7,10,9,8. Differences in height, gender, textures, materials, body proportions and anthropometry are readily apparent.}
\label{fig:subjects}
\end{figure*}

We used the open-source system Blender \cite{Blender} to render the motion of characters driven by the real motion capture data. Figure \ref{fig:pipeline}(d) shows a skeleton driven by the motion capture data. In order to generate plausible synthetic video sequences, a 3D character (i.e. avatar -- see Figure \ref{fig:pipeline}(e)) was created using the anthropometry of each volunteer. The open-source MakeHuman package \cite{MakeHuman} was chosen for this task. MakeHuman allows human-shaped templates to be formed into detailed 3D characters. Features like gender, age, muscle percentage, weight, height, body proportions, can be adjusted. Confounding and identity obscuring changes can also be applied to the visual appearance of characters, including clothing and hair.
A small sample of avatars, created for six different subjects from our dataset, is presented in Figure \ref{fig:subjects}; differences in height, gender, textures, materials, body proportions and anthropometry are readily apparent.


With MakeHuman, one can also attach the avatar of a human subject to a rig (i.e. a skeleton, known as armature), which then makes it possible to deform and move the character around. This is known as the \emph{skinning} procedure, in which each vertex of the 3D mesh model is associated with a certain bone in the rig. Each vertex is weighted differently, according to a physics engine that takes into account how the surface of the body moves relative to the skeleton.



In order to be able to deform the 3D avatar of a human subject (Figure \ref{fig:pipeline}(e)) according to the corresponding real motion capture data for that subject, the motion from the real motion capture data (Figure \ref{fig:pipeline}(d)) has to be ``transferred'' to the rig inside the 3D avatar (Figure \ref{fig:pipeline}(f)). This procedure is known as \emph{retargetting}. We found it necessary to adjust some inconsistencies between the representation of the skeleton obtained from the motion capture system and the representation of the skeleton using the MakeHuman armature. We provide examples of the processing required to do this in the supplementary material. Once done, it was possible to render the avatar corresponding to a given human subject within Blender, and to vary lighting, camera angles and several other parameters. Video sequences generated from avatar movements compared favourably with real video sequences captured from the corresponding human subjects.



\begin{figure*}[h]
\begin{center}
\bmvaHangBox{\fbox{\includegraphics[width=0.6\textwidth]{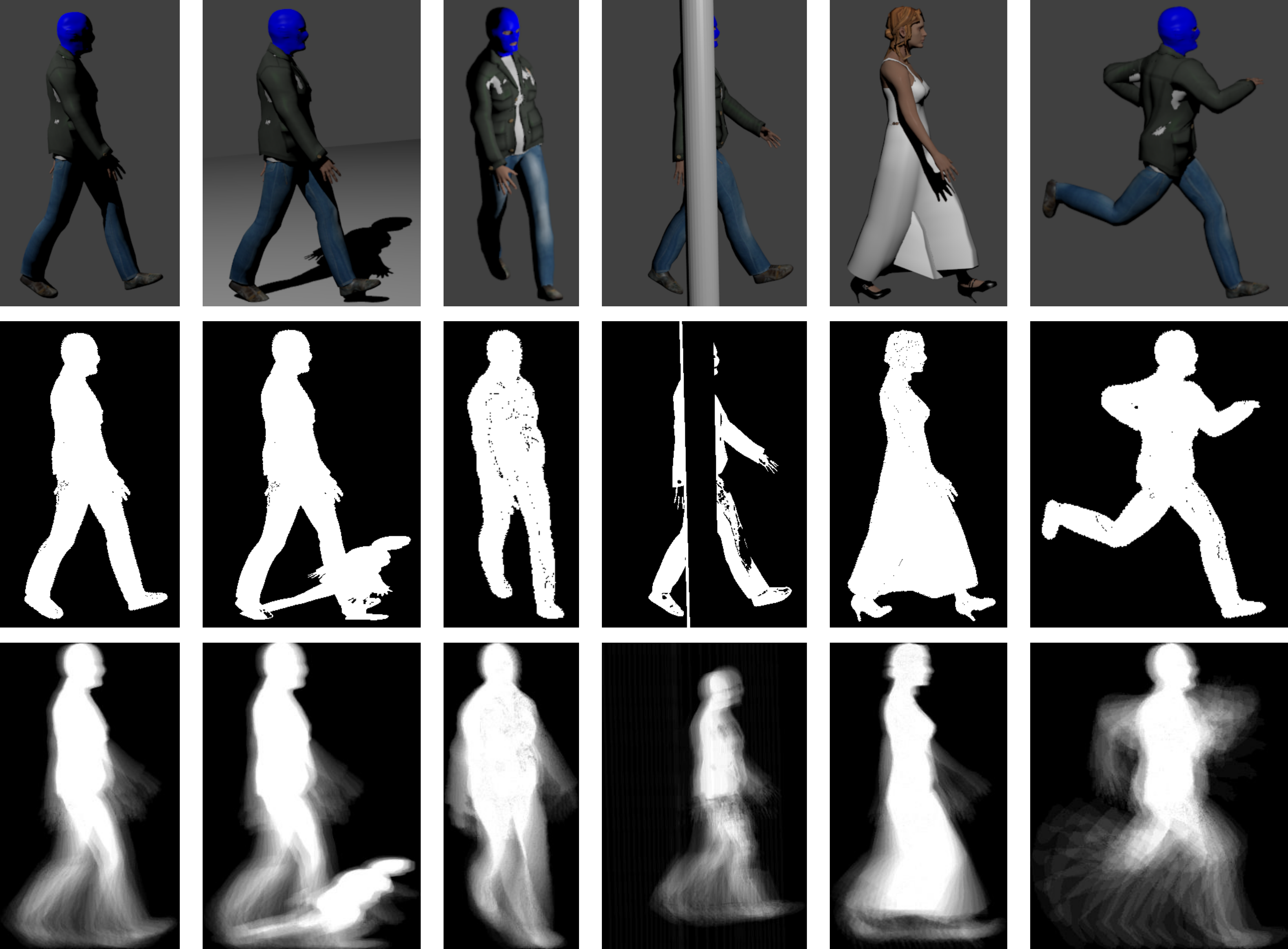}}}
\end{center}
   \caption{A small sample of the controllable confounding factors. Top row shows frames from the synthetic video data, middle row shows the corresponding binary silhouettes, and the bottom row shows the synthetic GEI features extracted for the corresponding gait cycles. It is readily apparent how these confounding factors directly affect the extracted gait features.}
\label{fig:covFactors}
\end{figure*}

The combination of real motion capture, data preparation, avatar construction and scripted rendering, allows synthetic frames to be generated with almost arbitrary degrees of variation. Samples of controllable confounding factors are presented in Figure \ref{fig:covFactors}, which shows frames from the synthetic video data, the corresponding binary silhouettes, and the GEI features calculated from the corresponding gait cycles. This data was used to create a large dataset across time, confounding factors and individuals. In addition to raw video and motion capture data from volunteers, the dataset contains the deformed 3D mesh representations for all subjects for a walking trial (5 km/h), allowing subject-specific and camera and pose-specific image sequences to be produced.


\section{Experiments}
\label{sec:experiments}


The process of data augmentation through motion capture and avatars is much more complex than that used in other forms of data augmentation in machine vision to date.  A key question that must be addressed is: to what extent is identity preserved in applying such a high degree of complexity? In order to evaluate the degree to which information on a subject's identity is passed from the real data onto the synthetic data, we chose to perform experiments using the Gait Energy Image (GEI) features \cite{Han}; these were used as gait features in order to train Support Vector Machines (SVM) to identify the subjects. GEI features are widely used for gait recognition, mainly because they are easy to extract. Due to employing the averaging procedure \cite{Han}, they are also relatively immune to errors such as foot sliding and jitter in human movement, as well as to noise present in the binary silhouettes due to suboptimal segmentation, which is very frequent in real-world applications.

The GEI feature requires a binary silhouette to be extracted from each frame in a video sequence, followed by an averaging operation across a complete gait cycle in time. The real binary sequences are calculated by converting all the frames from the real gait sequences into the LAB colour space and then subtracting the background from each frame in the video. The resulting grayscale image is then segmented by clustering the intensity values into two separate classes, corresponding to the foreground and the background pixel intensities. A simple thresholding procedure, according to that clustering, provides the real binary silhouettes. The segmentation of synthetic data is based on simple colour separation, which is facilitated by choosing a uniform colour for the background when rendering the scene.

Once the binary silhouettes are extracted, the video sequences are divided into gait cycles. Inspired by \cite{Sarkar}, we consider only the lower part of the human body and count the ``on'' (white) pixels for each frame in a video sequence. Assuming that when the legs are at the extreme positions (i.e. full stride stance) the number of ``on'' pixels reaches a maximum value, we plot that number for each frame in a video sequence. The result is a periodic signal with half the period of a gait cycle. In order to follow the correct definition of a gait cycle, one has to consider every second peak. The timings of these peaks define the boundaries of the gait cycles.

\begin{figure}[h]
\begin{tabular}{cc}
\bmvaHangBox{\fbox{\includegraphics[width=0.442\textwidth]{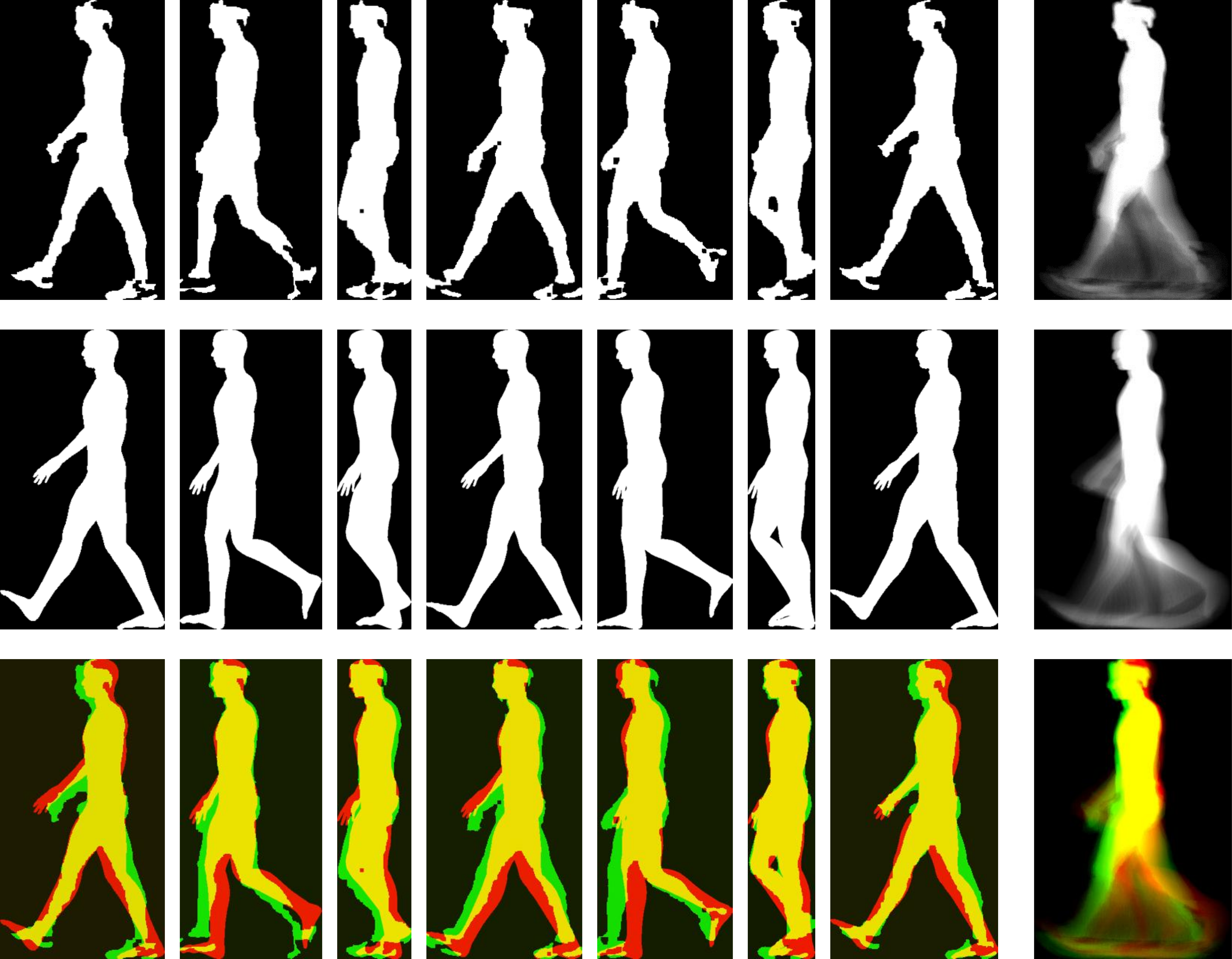}}}&
\bmvaHangBox{\fbox{\includegraphics[width=0.460\textwidth]{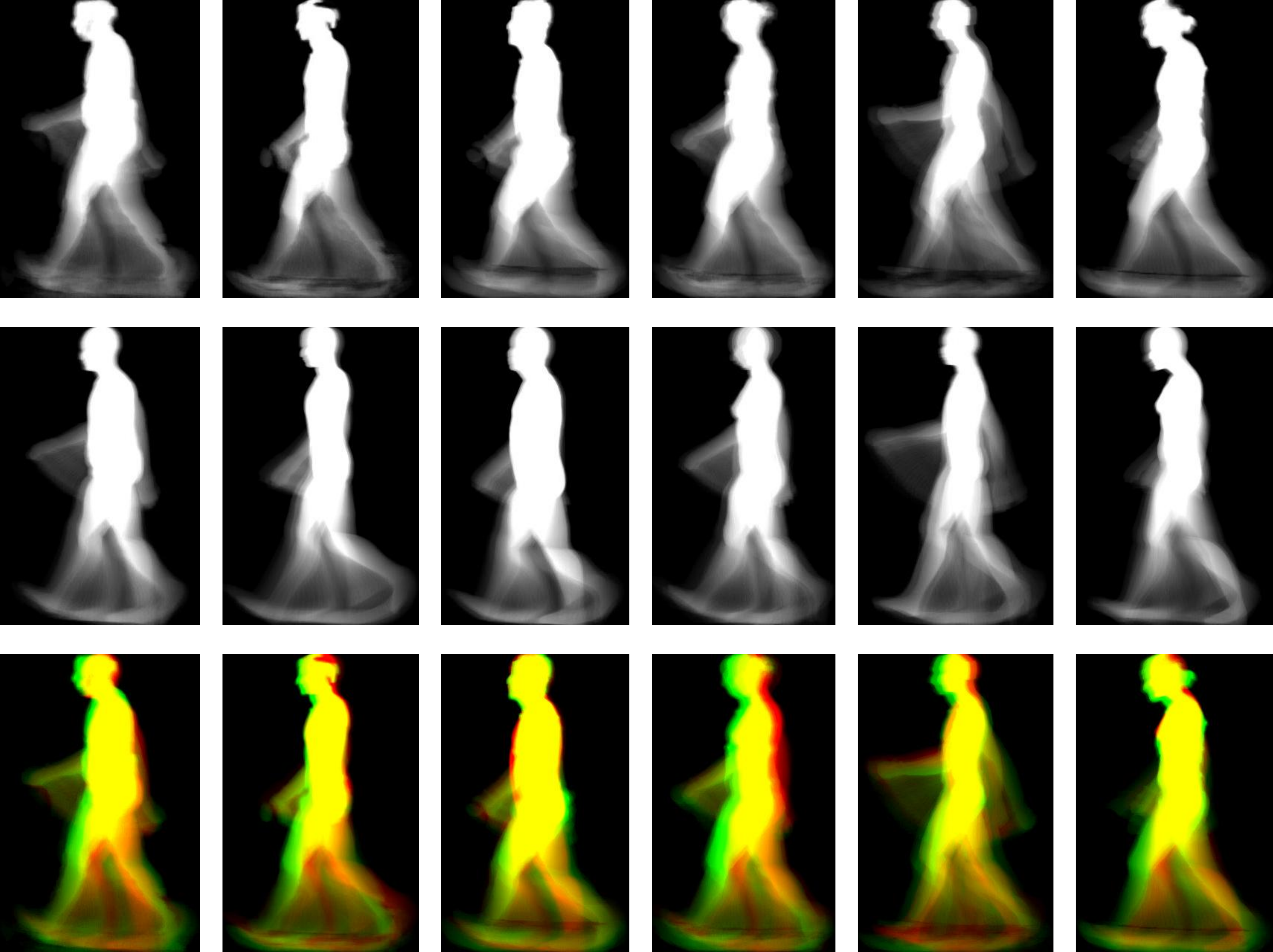}}}\\
(a)&(b)
\end{tabular}
   \caption{(a) Seven key frames of a gait cycle and the resulting GEI image, for subject 3, with the real binary silhouettes on top, the synthetic ones in the middle and the fusion of the two at the bottom (green for real, red for synthetic and yellow for their overlap). (b) GEI features for the subjects of Figure \ref{fig:subjects} (left to right, subjects 1,3,7,8,9 and 10), with the real GEIs on top, the synthetic ones in the middle and their fusion at the bottom (same colouring).}
\label{fig:realSynthGEIs}
\end{figure}

Having all the binary silhouettes extracted and resized down to 50$\times$30 pixels, we perform an average operation across frames for each gait cycle. This results in a normalised grayscale image, which is, by definition, the GEI image \cite{Han}. We apply further data augmentation by left-right flipping, systematically cropping the image features and then resizing back to 50$\times$30 \cite{Han}. Figure \ref{fig:realSynthGEIs}(a) presents examples of real (top row) and synthetic (middle row) binary silhouettes, and the resulting GEI images, for subject 3, as well as the fusion of them (bottom row). Figure \ref{fig:realSynthGEIs}(b) shows additional real and synthetic GEI feature images for the subjects presented in Figure \ref{fig:subjects}, along with the fusion of them.


Using the synthetic frames and corresponding GEIs, we conducted experiments designed to assess the level to which identity is preserved within the generated synthetic data sets. The first experiment compared binary silhouettes from real subjects with those from their avatars, driven by the real motion capture data. The second experiment compared the principal components of the 1500-dimensional GEIs; this is divided into two parts: a visualisation of the principal components of real and synthetic frames, and an exploration of training and testing recognition with both datasets separately and together (real/synthetic data augmentation).

%
%
%


\section{Results \& Discussion}
\label{sec:resultsAndDiscussion}


In order to assess the degree of agreement between the real and synthetic binary silhouettes, we calculated the Jaccard index between them, for each frame, for each individual, with frame alignment. The Jaccard index is defined as the size of the intersection divided by the size of the union of the two regions (see equation \ref{eq:JaccardIndex}), and approaches 1 for perfect alignment.

\begin{equation}
	J(A,b) = \frac{| A \bigcap B |}{| A \bigcup B |} = \frac{| A \bigcap B |}{| A | + | B | - | A \bigcap B |}
	\label{eq:JaccardIndex}
\end{equation}

\begin{figure}[ht]
\begin{center}
\bmvaHangBox{\fbox{\includegraphics[width=0.85\textwidth]{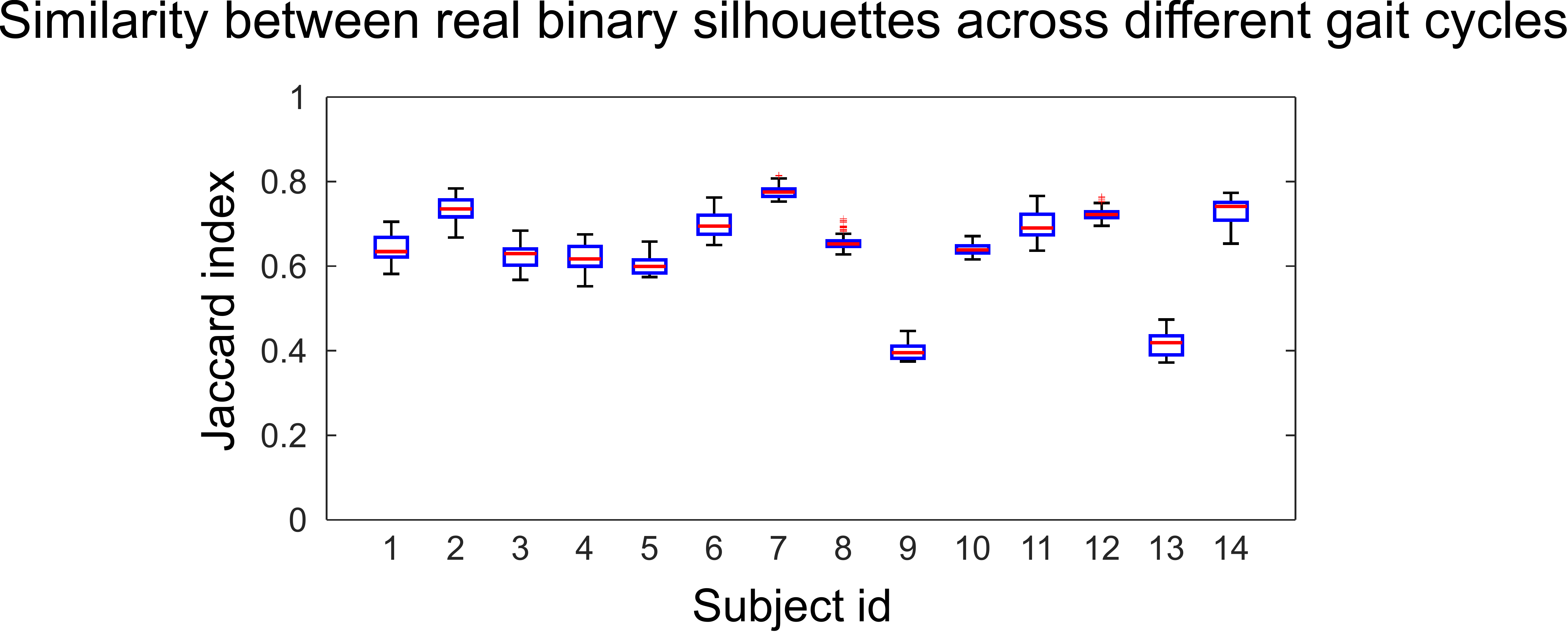}}}\\
\end{center}
   \caption{Similarity between real binary silhouettes, across different gait cycles of the same Subject, for a walking trial, measured using the Jaccard index between the corresponding binary silhouettes (a baseline for the similarity between real and synthetic binary silhouettes).}
\label{fig:JIs_boxplot_RealVsReal}
\end{figure}

\begin{figure}[ht]
\begin{center}
\bmvaHangBox{\fbox{\includegraphics[width=0.85\textwidth]{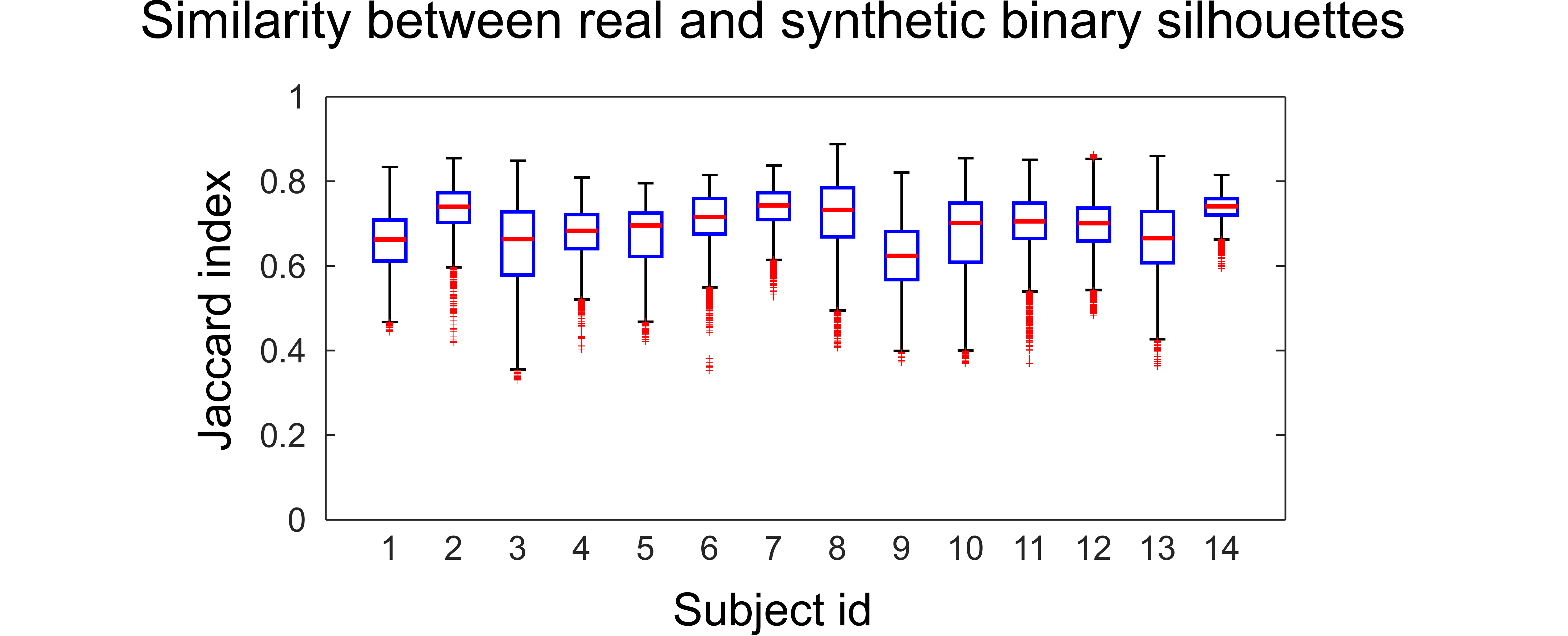}}}\\
\end{center}
   \caption{Similarity between real and synthetic binary silhouettes for a walking trial of Subjects 1-14, measured using the Jaccard index between the corresponding binary silhouettes.}
\label{fig:JIs_boxplot_RealVsSynthetic}
\end{figure}

We also calculated the similarity between real binary silhouettes of the same subject, across different gait cycles, which serves as a baseline (see Figure \ref{fig:JIs_boxplot_RealVsReal}). This can be seen as the realistic upper bound for the Jaccard index, when used as a metric to assess the consistency of silhouette alignment across cycles. The calculated similarity using the Jaccard index over the entire walking trial (at 5 km/h) for subjects 1-14 is presented in Figure \ref{fig:JIs_boxplot_RealVsSynthetic}. With Figure \ref{fig:JIs_boxplot_RealVsReal} as a baseline, it shows that synthetically generated data aligns well with real sequences.


There are several reasons why real and synthetic binary silhouettes do not fully match. Hairstyle, for example, can play a significant role in the appearance of a subject. When not properly modelled, it may affect the match to a great extent. Also, the Jaccard index is neither rotation-invariant (a possible roll angle offset on the video camera capturing the real data may cause this misalignment) nor translation-invariant, which means that small miss-alignments -- see Figure \ref{fig:realSynthGEIs}(a) -- decrease the similarity values dramatically. Since we compare silhouettes from single frames with the Jaccard index, noise introduced by segmentation of real image sequences -- see Figure \ref{fig:realSynthGEIs}(a) (top row) -- shows common imperfections which are not present in the generated synthetic data. This decreases the average Jaccard index values reported.

\begin{figure}[h]
\begin{tabular}{cc}
\bmvaHangBox{\fbox{\includegraphics[width=0.43\textwidth]{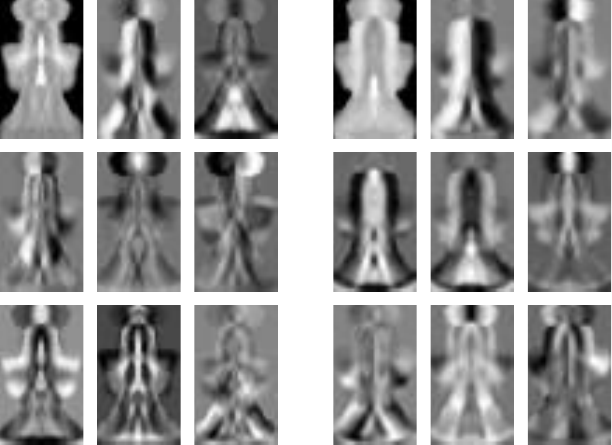}}}&
\bmvaHangBox{\fbox{\includegraphics[width=0.463\textwidth]{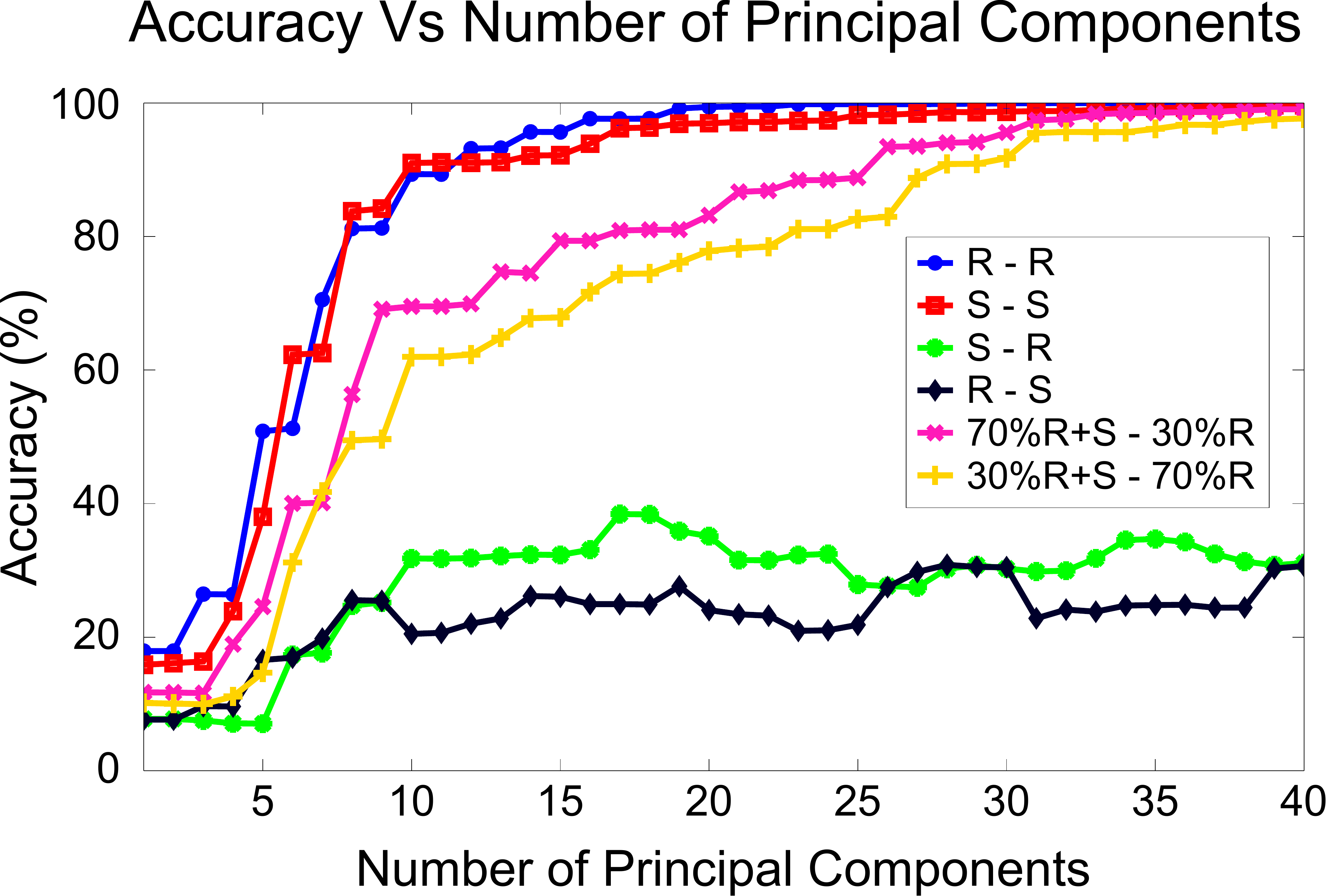}}}\\
(a)&(b)
\end{tabular}
\caption{(a) The first nine principal components calculated on the GEI features, using real data (left) and synthetic data (right); it is readily apparent that the feature space in the two cases is similar but not identical (b) results from our experiments, with different training and testing sets, using combinations of real and/or synthetic data. We present the accuracies achieved, with the GEI features being projected to different number of principal components.}
\label{fig:results}
\end{figure}

However, as seen in Figure \ref{fig:realSynthGEIs}, qualitatively, the synthetic binary silhouettes keep a subject's individual locomotion style. This is also readily apparent, quantitatively, in Figure \ref{fig:results}(b), where the accuracy for the experiments in which training and testing data were solely synthetic, is relatively high. This means that some discriminating characteristics containing a subject's identity is preserved and passed onto the generated synthetic data, making them suitable for data augmentation.
 
Results from our experiments -- presented in Figure \ref{fig:results}(b) -- suggest that information about the identity of the subjects is retained within the synthetically generated data. Figure \ref{fig:results}(a) shows the first nine principal components calculated on the GEI features, using real data (left) and synthetic data (right), whereas Figure \ref{fig:results}(b) shows results of our experiments. Under same-mode conditions (either training and testing with solely real data \textbf{R -- R}, or training and testing with solely synthetic data \textbf{S -- S}) we achieve the best results. It is also readily apparent that, given a sufficient number of principal components, using a limited number of real data, augmented with the synthetically generated data (cases \textbf{70\%R+S -- 30\%R} and \textbf{30\%R+S -- 70\%R}), results in identification of the subjects with an accuracy of more than 95\%.  However, in the cross-mode conditions (training with real data and testing with synthetic data \textbf{R -- S}, or the other way around \textbf{S -- R}) we get the lowest results. This is probably because the features learnt solely by real data and solely by synthetically generated data are similar, but not necessarily identical (see Figure \ref{fig:results}(a)); therefore, the feature space learnt is not actually shared, but rather shifted between the two cases.

The use of a treadmill allowed us to limit the area of motion capture, while being able to capture long sequences of the order of several minutes. Whilst possibly altering natural gait, the use of a treadmill is supported by a study of the biomechanics of human motion \cite{Lee}, which compared overground with treadmill walking in healthy individuals. The differences between overground and treadmill walking were found to be surprisingly small.

Finally, because anthropomorphic data was introduced into the avatars by hand, details of body appearance have not been fully incorporated in the 3D mesh models. Since subsequent steps, including silhouette calculations, depend on these meshes, there may be loss of detail introduced by this manual process. This loss of detail may also be the result of the kinematics of the gait not being perfectly transferred to the avatar, due to physics deformation of the avatar differing from that of the real subject.

\section{Conclusion and Future Work}
\label{sec:conclusion}


We suggested a method to generate synthetic video data for data augmentation of gait sequences. The process allows sequences to be generated with multiple confounding factors, and allows exploratory work into training machine/deep learning algorithms for fully-invariant gait recognition, with a far greater amount of synthetic training and test data than would otherwise be possible.

A multi-modal dataset produced during this study is based on real motion capture data, with a total number of more than 6.5 million frames. Augmenting the dataset with synthetic data introduces further confounding factors, allowing even larger amounts of data to be synthesized. Even though the amount of synthetically generated data is limited by the amount of real mocap data collected, the proposed methodology attempts to maximise that number by modelling as many of the confounding factors stated in Section \ref{sec:intro} as possible. For example, introducing different subject-camera relative viewing angles, with a simple sampling of the angular space with a step of $5^\circ$, in both longitudinal and latitudinal directions, results in 703 different pairs of azimuth and elevation angles. Augmenting the data by introducing a single confounding factor like subject-camera relative viewing angle, increases the number of frames to more than 4.5 billion.

As seen qualitatively in Figure \ref{fig:realSynthGEIs} and quantitatively in Figure \ref{fig:results}(b), the synthetic video data contains identity information that discriminates subjects. One could argue that using avatars driven by motion capture to generate subject-specific image data is a poor substitute to using real image data for training systems in identity recognition. However, there are still open questions about the best way to attain invariant recognition of human gait, given the very large number of confounding factors that affect the performance of recognition. By generating a large set of data that -- at least visually -- replicates some of the variability in appearance of subjects, it is possible to explore features and techniques that are more robust than existing gait recognition methods.

Future work includes a) expanding the span of confounding factors in the datasets (age, gender, body shape, physical fitness); b) improving confounding factors, such as camera models and garment deformations, to close the gap between synthetic and real data; c) continuing the collection of motion capture data from additional subjects to increase the population of our dataset. The dataset, along with the simulation files, is publicly available at \url{http://www.bicv.org/datasets/}.

\bibliography{cccbib}
\end{document}